\documentclass[a4paper,twoside,french]{article}
\usepackage{rfia2000}
\usepackage[T1]{fontenc}
\usepackage{babel}
\usepackage{times}
\usepackage{graphicx}
\usepackage{tikz}
\usepackage{pgfplots}
\usepackage[hidelinks]{hyperref}
\usepackage[utf8]{inputenc}

\begin{document}
\date{}
\title{\Large\bf Histogram of Oriented Depth Gradients for Action Recognition}
\author{\begin{tabular}[t]{c@{\extracolsep{8em}}c}
Nachwa~ABOU BAKR${}^1$  & James~CROWLEY${}^2$ \\
\end{tabular}
{} \\
 \\
        Univ. Grenoble Alpes, INRIA, Grenoble INP,  CNRS, Laboratoire LIG  \\
${}^1$ nachwa.aboubakr@inria.fr \\
${}^2$ james.crowley@inria.fr
}
\maketitle
\thispagestyle{empty}
\subsection*{R\'esum\'e}
{\em
Cet article présente des expériences menées sur la perception de la profondeur de mouvements dans le cadre de la reconnaissance visuelle d'actions à partir de mesures locales extraites de séquences vidéo RGBD encodées selon la norme MPEG. 
Nous démontrons que ces mesures peuvent être combinées avec des descripteurs RGB spatio-temporels locaux pour l'élaboration d'une méthode de calcul efficace pour la reconnaissance d'actions. 
Des vecteurs de Fisher sont utilisés pour encoder et concaténer un descripteur de profondeur avec les descripteurs existants RGB locaux. Ces vecteurs sont ensuite utilisés pour la reconnaissance d'actions manuelles à l'aide d'un classifieur linéaire SVM.
Nous comparons l'efficacité de ces mesures à l'état de l'art sur deux jeux de données récents pour la reconnaissance d'actions dans des environnements culinaires.
}
\subsection*{Mots Clef}
Vision, Profondeur, Reconnaissance d'Actions, Codage MPEG.

\subsection*{Abstract}
{\em
In this paper, we report on experiments with the use of local measures for depth motion for  visual action recognition from MPEG encoded RGBD video sequences. We show that such measures can be combined with local space-time video descriptors for appearance to provide a computationally efficient method for recognition of actions. 
Fisher vectors are used for encoding and concatenating a depth descriptor with existing RGB local descriptors. 
We then employ a linear SVM for recognizing manipulation actions using such vectors.
We evaluate the effectiveness of such measures by comparison to the state-of-the-art using two recent datasets for action recognition in kitchen environments.
}
\subsection*{Keywords}
Vision, Depth, Action Recognition, MPEG encoding.

\section{Introduction}
Visual action recognition is the process of labelling image sequences with action labels.  
In this work we investigate the use of dense local descriptors for action recognition.
Dense sampling has been shown to improve results when compared to sparse interest points for action recognition \cite{SOA:wanglocalevaluation}. 
Requirements for reduced computational time have lead some authors to investigate methods that extract motion vectors directly from compressed video signals.  For example, methods have been proposed to extract flow directly from MPEG encoded video \cite{SOA:MFmethod}. The resulting method has been shown to provide useful description of Dense Trajectories (DT) \cite{SOA:DTmethod}, at the cost of some loss of recognition accuracy.
This has led us to ask if this approach can be extended to provide fast computation of dense local descriptors from the depth channel of MPEG encoded RGBD image sequences.\par
In the following, we report the results on experiments in recognition of action labels using features extracted from depth channel of Kinect-like sensors using MPEG encoded motion flow. We describe a new descriptor based on oriented gradients of depth extracted from the motion channel of RGBD image sequences. This descriptor is used in addition to the MPEG Flow (MF) descriptors based on the RGB channels from the same sequence. 
Local video descriptors using Dense Trajectories \cite{SOA:DTmethod} and MPEG Flow \cite{SOA:MFmethod} are employed as baseline descriptors for comparison.  
 
DT, as a video descriptor for action recognition, starts by sampling interest points densely at regular fixed grids. Detected interest points are then tracked over time and points for which no motion is detected are  eliminated. This set of trajectories are then used as space-time interest points and  descriptors are computed around those points. Four descriptors have been used in DT: Histogram of Oriented Gaussian (HOG), Histogram of Optical Flow (HOF), Motion Boundary Histogram in x,y directions (MBH$_x$, MBH$_y$).
An analysis on computational requirements of DT shows that 61\% of the running time is spent on the computation of optical flows. Reducing this computational burden is an interesting challenge. 

The MF method exploits the estimated motion vectors used for video compression and uses them as interest points. This change provides a significant improvement in the time required for extraction of features at the cost of a small decrease in recognition accuracy \cite{SOA:MFmethod}.

Both DT and MF are video descriptors computed around motion trajectories. The primary differences between these two methods are that the interest points to be described in MF are at sparse positions of motion fields. In DT, points are sampled at the positions of a regular fixed grid, and they are tracked over time to build trajectories. 
Motion is encoded in DT using Gunnar-Farneb\"ack optical flows while MF uses estimated motion fields during video compression. Our goal is to determine if it is possible to improve MF results by exploiting data from depth channel of Kinect-like sensors. 


\section{Histogram of Oriented Depth Gradients}
Depth can capture the structural features of objects. In addition to encoding the position of objects in space. Thus it is reasonable to ask if we can improve the results of MF video descriptor by including the depth channel to better describe the scene. To do so, we compute gradients of depth which encodes the structure of the depth image. As in MF, interest points are a sparse set of points extracted while compressing video. This ensures describing area around motion points.\\
Practically, a volume of $32 \times 32 \times 15$ is computed around interest points. This volume is divided by a grid of $2 \time 2 \time 3$ which produces a grid cell of $16 \time 16 \times 5$. For each grid cell, we extract the gradients by applying the Sobel filter $[1,0,-1]$ in a $16 \times 16$ pixels. Then, an 8-orientation bins discretized histogram is built to encode the gradient orientations. Gradient magnitudes are used to weight the gradient orientations. For each temporal slice the gradient is normalized using $l2$-normalization.\\  
%
After that descriptors are encoded by transforming the collection of local video descriptors \{$x_i, \dots, x_N$\} into a fixed-size vector representation. Following \cite{SOA:MFmethod}, we chose to use Fisher Vectors (FV) \cite{FisherVectors} for descriptor encoding. In particular, only first-order differences of FV are used. 
Action categories are then classified by a linear one-vs-rest SVM classifier.

\section{Experiments}
%
Experiments are performed on two publicly available datasets: 50 salads \cite{Datasets:50salads} and Actions of Cooking Eggs (ACE) \cite{Dataset:ACE}. 
For 50 salads dataset; we used provided annotations of \textit{core} actions only as verbs without taking into account their corresponding objects; in total, we used 6 different action labels. ACE also provides 6 different annotated actions but only one level of granularity which includes pre- and post movements of performed actions. For the evaluation, we measured the mean of Average Precision (mAP).\\
We consider two image sequences; one provides an RGB view of the scene and the other encodes scene depth. To study the impact of motion depth image sequences in describing actions, we extracted a Histogram of Oriented Depth Gradients (HODG) from depth image sequences. We extracted as well RGB descriptors of DT \footnote{\url{https://lear.inrialpes.fr/people/wang/dense_trajectories}} and MF \footnote{\url{https://github.com/vadimkantorov/cvpr2014}}.
Combining both descriptors improves the results of MF method on two datasets as shown in Table \ref{tab:results_all} where RGB+HODG is a concatenation of encoded descriptors by FV. We compare the results to DT since MF follows the same structure and set of descriptors.
\par
DT approach achieved the best results in this experiment at the cost of the computation time. In the other hand, adding HODG to MF descriptors achieves better results than \cite{SOA:MFmethod} while preserving the computational cost \footnote{Each channel of the sensor can be treated individually in the descriptor extraction process}. Table \ref{tab:timeinfps} shows the number of processed frames per second while computing both RGB-based and HODG using the implementation of DT and MF. From the tables, we can also see that HODG alone performs relatively good in action recognition.

\begin{table}[!t]
\centering
\begin{tabular}{l|cc} \hline
         	 & 50salads & ACE     \\ \hline
DT(RGB) 	 & \textbf{99.60}\%  & 94.12\% \\
MF(RGB)  	 & 90.56\%  & 92.04\% \\
MF(HODG)	 & 86.29\%  & 89.94\% \\
MF(RGB+HODG) & 94.73\%  & \textbf{99.15}\% \\ \hline
\end{tabular}
\caption{The results of our experiments using mAP.}
\label{tab:results_all}
\end{table}
\begin{table}[!t]
\centering
\begin{tabular}{l|cc} \hline
      				  & DT (fps)  & MF (fps) \\ \hline
RGB (HOG, HOF, MBH)   & 3.87 	  & 27.78  	 \\
Depth (HODG) 	  & 6.41 	  & 111.13 	 \\ \hline
\end{tabular}
\caption{The number of processed frames per second while descriptors extraction for 50 salads dataset.}
\label{tab:timeinfps}
\end{table}
\section{Conclusion}
In this work, we reported an improvement of the results of MF method by using the benefit of depth channel in Kinect-like sensors. Since depth channels provide separate images that can be treated in parallel with no additional cost or sequentially while preserving a close to real-time capability. 
The idea is to combine the gradients of depth image sequences (HODG) extracted at sparse motion fields which are estimated in video compression to the set of RGB descriptors of MF \cite{SOA:MFmethod}. Both RGB and Depth descriptors are encoded using FV. 
\par
We found that trajectories of depth image sequences, described by HODG, can describe manipulation actions with an acceptable recognition precision. 
To this end, we evaluated our proposed method on two recent datasets; 50 salads, and ACE. We recorded the computational time during extraction of descriptors using both methods. Results show a significant improvement in recognition precision with almost no significant additional cost. 


\end{document}